\documentclass{article}
\usepackage{microtype}
\usepackage{graphicx}
\usepackage{subfigure}
\usepackage{pdfpages}
\usepackage{hyperref}
\usepackage{url}

\usepackage{array}
\usepackage{graphicx}

\usepackage{amsmath,amssymb}
\usepackage{nccmath}
\usepackage{color}
\usepackage{multirow}
\usepackage{tabularx, booktabs}

\newcolumntype{L}[1]{>{\raggedright\let\newline\\\arraybackslash\hspace{0pt}}m{#1}}
\newcolumntype{C}[1]{>{\centering\let\newline\\\arraybackslash\hspace{0pt}}m{#1}}
\newcolumntype{R}[1]{>{\raggedleft\let\newline\\\arraybackslash\hspace{0pt}}m{#1}}
\newcolumntype{x}[1]{>{\centering\arraybackslash\hspace{0pt}}p{#1}}

\usepackage[accepted]{icml2018}
\usepackage{caption}
\icmltitlerunning{WSNet: Compact and Efficient Networks Through Weight Sampling}
\begin{document}

\twocolumn[
\icmltitle{WSNet: Compact and Efficient Networks Through Weight Sampling}


\begin{icmlauthorlist}
\icmlauthor{Xiaojie Jin}{nus,snap}
\icmlauthor{Yingzhen Yang}{snap}
\icmlauthor{Ning Xu}{snap}
\icmlauthor{Jianchao Yang}{bd}
\icmlauthor{Nebojsa Jojic}{msr}
\icmlauthor{Jiashi Feng}{nus}
\icmlauthor{Shuicheng Yan}{360,nus}
\end{icmlauthorlist}

\icmlaffiliation{snap}{Snap Inc. Research, Los Angeles, USA}
\icmlaffiliation{nus}{National University of Singapore, Singapore}
\icmlaffiliation{bd}{Bytedance Inc., Menlo Park, USA}
\icmlaffiliation{msr}{Microsoft Research, Redmond, USA}
\icmlaffiliation{360}{360 AI Institute, Beijing, China}

\icmlcorrespondingauthor{Xiaojie Jin}{xjjin0731@gmail.com}
\icmlkeywords{Deep Learning, Model compression and acceleration}
\vskip 0.3in
]

\printAffiliationsAndNotice{}

\begin{abstract}
We present a new approach and a novel architecture, termed WSNet, for learning compact and efficient deep neural networks. Existing approaches conventionally learn full model parameters independently and then compress them via \emph{ad hoc} processing such as model pruning or filter factorization. Alternatively, WSNet proposes learning model parameters by sampling from a compact set of learnable parameters, which naturally enforces {parameter sharing} throughout the learning process. We demonstrate that such a novel weight sampling approach (and induced WSNet) promotes both weights and computation sharing favorably. By employing this method, we can more efficiently learn much smaller networks with competitive performance compared to baseline networks with equal numbers of convolution filters. Specifically, we consider learning compact and efficient 1D convolutional neural networks for audio classification. Extensive experiments on multiple audio classification datasets verify the effectiveness of WSNet. Combined with weight quantization, the resulted models are up to \textbf{180$\times$} smaller and theoretically up to \textbf{16$\times$} faster than the well-established baselines, without noticeable performance drop.
\end{abstract}

\section{Introduction}
\label{sec:intro}
Despite remarkable successes in various applications, deep neural networks (DNNs) usually suffer following two problems that stem from their inherent huge parameter space. First, most of state-of-the-art deep architectures are prone to over-fitting even when trained on large datasets~\citep{vgg,googlenet}. Secondly, DNNs usually consume large amount of storage memory and energy~\citep{deep-compression}, which makes it difficult to use them in devices with limited memory and power (such as portable devices or chips). Different from most existing works~\citep{han2015learning,li2016pruning,jaderberg2014speeding,lebedev2014speeding,dark-knowledge} on model compression and acceleration that ignore the strong dependencies among weights and learn filters independently based on existing network architectures, this paper proposes to explicitly enforce the parameter sharing among filters to more effectively learn compact and efficient deep networks.

In this paper, we propose a \textbf{W}eight \textbf{S}ampling deep neural network (\textit{i.e.}\ WSNet) to significantly reduce both the model size and computation cost, achieving more than 100$\times$ smaller size and up to 16$\times$ speedup at negligible  performance drop or  even achieving better performance than the baseline ({\textit{i.e.} conventional networks that learn  filters independently}). Specifically, WSNet is parameterized by  layer-wise \emph{condensed filters} from which each filter participating in actual convolutions can be directly sampled, in both spatial and channel dimensions. Since  condensed filters have significantly fewer parameters than  independently trained filters as in conventional CNNs, learning by sampling from them makes WSNet a more compact model compared to conventional CNNs.
In addition, to reduce the ubiquitous computational redundancy in convolving the overlapped filters and input patches, we propose an integral image based method to dramatically reduce the computation cost of WSNet in both training and inference. The integral image method is also advantageous because it enables weight sampling with different filter size and minimizes computational overhead to enhance the learning capability of WSNet.

In order to demonstrate the efficacy of WSNet, we conduct extensive experiments on challenging audio classification tasks. On each test dataset, including ESC-50 \citep{piczak15}, UrbanSound8K \citep{Salamon14}, DCASE \citep{Stowell15} and MusicDet200K (a self-collected dataset, as detailed in Section~\ref{sec:exp}), WSNet significantly reduces the model size of the baseline by 100$\times$ with comparable or even higher classification accuracy. When compressing more than 180$\times$, WSNet is only subject to negligible accuracy drop. {At the same time, WSNet significantly reduces the computation cost (up to 16$\times$).} Such results strongly establish the capability of WSNet to learn compact and efficient networks. Last but not the least, we provide an intuitive method to extend WSNet from 1D CNNs to 2D CNNs. Experimental results on MNIST and CIFAR10 strongly evidence the potential capability of WSNet to learn efficient networks on 2D CNNs.

\section{Related Works}

\subsection{Deep Model Compression and Acceleration}
Recent works in network compression adopt weight pruning~\citep{han2015learning,collins2014memory,anwar2015structured,lebedev2015fast,kim2015compression,thinet,li2016pruning}, filter decomposition~\citep{sindhwani2015structured,denton2014exploiting,jaderberg2014speeding}, hashed networks~\citep{hashnet,freshnet} and weight quantization~\citep{deep-compression}. However, although those works reduce model size, they also suffer from large performance drop. \citet{st1} and \citet{st2} are based on student-teacher approches which may be difficult to apply in new tasks since they require training a teacher network in advance. \citet{denil2013predicting} predicts parameters based on a few number of weight values. \citet{iht} proposes an iterative hard thresholding method, but only achieve relatively small compression ratios. \citet{gong2014compressing} uses a  binning method which can only be applied over fully connected layers. ~\citet{dark-knowledge} compresses deep models by transferring the knowledge from pre-trained larger networks to smaller networks.

	In terms of deep model acceleration, the factorization and quantization methods listed above can reduce computation latency in inference.
FFT~\citep{mathieu2013fast} and LCNN~\citep{bagherinezhad2016lcnn} are also used to speed up computation in pratice. Comparatively, WSNet is superior because it learns networks that have both smaller model size and faster computation versus baselines.
\subsection{Efficient Model Design}
WSNet presents a class of novel models with the appealing properties of a small model size and small computation cost. Some recently proposed efficient model architectures include the class of Inception models~\citep{googlenet,batchnorm,xception}, the class of Residual models~\citep{residual,resnext,dpn} and the factorized networks which use fully factorized convolutions. MobileNet~\citep{mobilenet} and Flattened networks~\citep{flattenet} are based on factorization convolutions. ShuffleNet~\citep{shufflenet} uses group convolution and channel shuffle to reduce computational cost. {Compared with the above works, WSNet presents a new model design strategy which is more flexible and generalizable: the parameters in deep networks can be obtained conveniently from a more compact representation through the proposed weight sampling method.}

	\subsection{Audio classification}
	Audio classification aims to classify the surrounding environment where an audio stream is generated given the audio input~\citep{Barchiesi15}. 
Compared with other CNN based methods which use pre-computed features, \textit{e.g.} MFCC \citep{Pols66,Davis80} and spectrogram \citep{Flanagan72}, recently proposed SoundNet \citep{soundnet} yields significant the state-of-the-art results by directly taking one dimensional raw wave signals as input. In this paper, we demonstrate that the proposed WSNet achieves a comparable or even better performance than SoundNet at a significantly smaller size and faster speed.
	\section{Method}\label{sec:method}
        \subsection{Notations} 
        \label{sec:notation}
	Before diving into the details, we first introduce the notations used in this paper. The traditional 1D convolution layer takes as input the feature map $\mathbf{F} \in \mathbb{R}^{T\times M}$ and produces an output feature map $\mathbf{G} \in \mathbb{R}^{T \times N}$ where $(T,M,N)$ denotes the spatial length of input, the channel of input and the number of filters respectively. We assume that the output has the same spatial size as input which holds true by using zero padded convolution. The 1D convolution kernel $\mathbf{K}$ used in the actual convolution of WSNet has the shape of $(L, M, N)$ where $L$ is the kernel size.  Let $\mathbf{k}_n, n\in \{1,\cdots N\}$ denotes a filter and $\mathbf{f}_t, t\in \{1,\cdots T\}$ denotes a input patch that spatially spans from $t$ to $t+L-1$, then the convolution assuming stride one and zero padding is computed as:

	\begin{equation} \label{eq:conv}
	\mathbf{G}_{t,n}  = \mathbf{f}_t \cdot \mathbf{k}_n
	= \sum_{l=0}^{L-1}\sum_{m=0}^{M-1}\mathbf{F}_{t+l,m} \times \mathbf{K}_{l,m,n},
	\end{equation}
	where   $\cdot$ stands for the vector inner product. Note we omit the element-wise activation function to simplify the notation.

	In WSNet, instead of learning each weight independently, $\mathbf{K}$ is obtained by sampling from a learned \textit{condensed filter} $\mathbf{\mathbf{\Phi}}$ which has the shape of $(L^*, M^*)$. The goal of training WSNet is thus cast to learn more compact DNNs which satisfy the condition of $L^*M^* < LMN$. WSNet uses a condensed filter per convolutional layer. To quantize the advantage of WSNet in achieving compact networks, we  define the \textit{compactness} of $\mathbf{K}$ in a learned layer in WSNet w.r.t. the conventional layer with independently learned weights as:
	\begin{equation}\label{eq:compactness}
	\text{compactness} = \frac{LMN}{L^*M^*}.
	\end{equation}
	In the following section, we demonstrate WSNet learn compact networks by sampling weights in two dimensions: the spatial dimension and the channel dimension.
	\begin{figure}
		\centering
		\includegraphics[width=\linewidth]{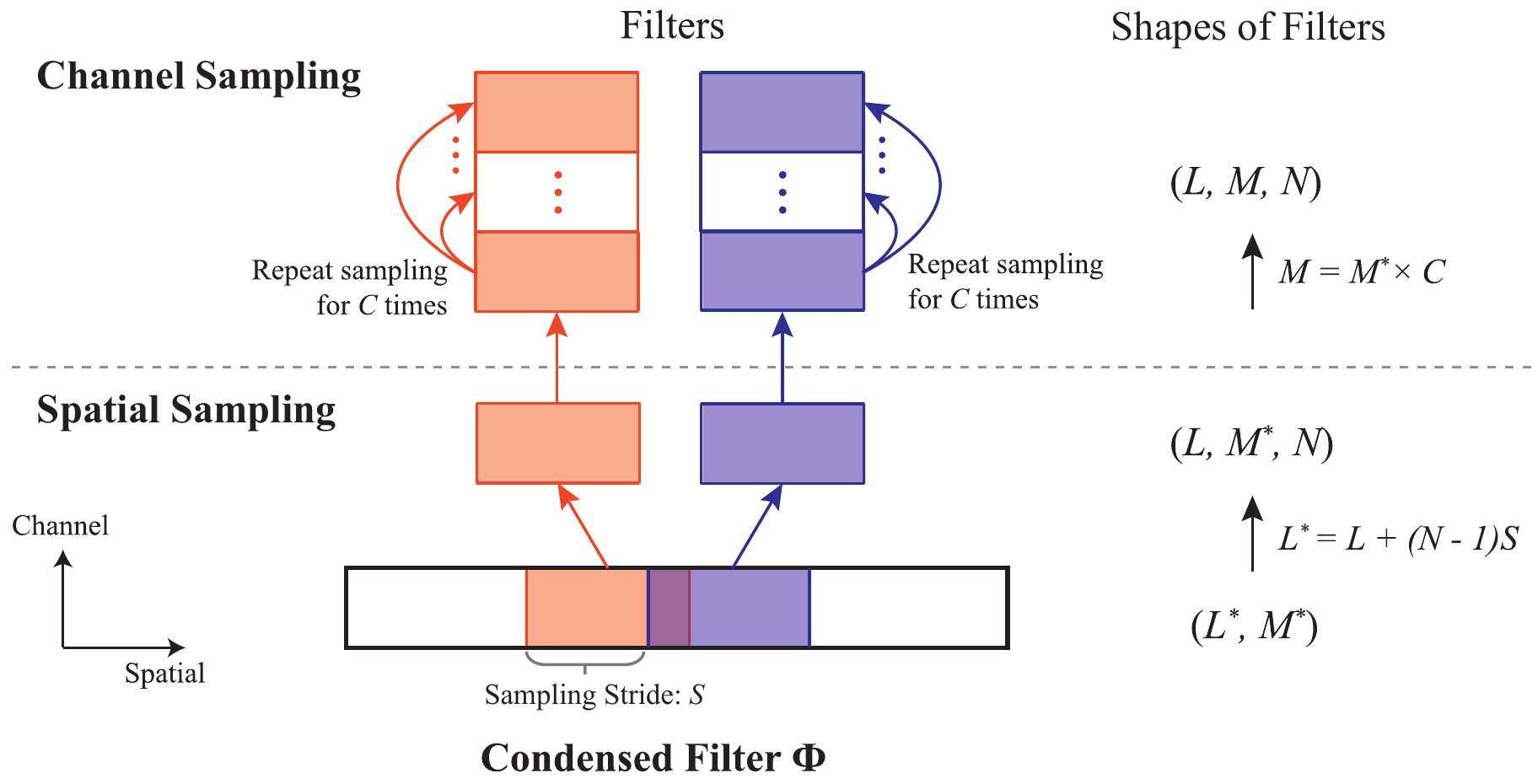}
		\caption{Illustration of WSNet that learns small condensed filters with weight sampling along two dimensions: spatial dimension (the bottom panel) and channel dimension (the top panel). The figure depicts  procedure of generating two continuous filters (in pink and purple respectively) that convolve with input. In \textbf{spatial sampling}, filters are extracted from the condensed filter with a stride of $S$. In \textbf{channel sampling}, the channel of each filter is sampled repeatedly for $C$ times to achieve equal with the input channel. Please refer to Section~\ref{sec:weightsampling} for detailed explanations.  All figures in this paper are best viewed in zoomed-in pdf.}
		\label{fig:illustration}
	\end{figure}

	\subsection{Weight sampling} \label{sec:weightsampling}
	\subsubsection{Along spatial dimension}\label{sec:spatial-sampling}
	In conventional CNNs, the filters in a layer are learned independently which presents two disadvantages. Firstly, the resulted DNNs have a large number of parameters, which impedes their deployment in computation resource constrained platforms. Second, such over-parameterization makes the network  prone to overfitting and getting stuck in (extra introduced) local minima. To solve these two problems, a novel weight sampling method is proposed to efficiently reuse the weights among filters. Specifically, in each convolutional layer of WSNet, all convolutional filters $\mathbf{K}$ are sampled from the condensed filter $\mathbf{\Phi}$, as illustrated in Figure~\ref{fig:illustration}. By scanning the weight sharing filter with a window size of $L$ and stride of $S$, we could sample out $N$ filters with filter size of $L$. Formally, the equation between the filter size of the condensed filter and the sampled filters is:
	\begin{equation}
	\label{eq:s_comp}
	L^* = L + (N - 1) S.
	\end{equation}
	The \textit{compactness} along spatial dimension is $\frac{L M^*N }{L^*M^*} \approx \frac{L}{S}$. Note that since the minimal value of $S$ is 1, the minimal value of $L^*$ (\textit{i.e.} the minimum spatial length of the condensed filter) is $L+ N - 1$ and the maximal achievable compactness is therefore $L$.
	\subsubsection{Along Channel dimension} \label{sec:channel-sampling}
	Although it is experimentally verified that the weight sampling strategy could learn compact deep models with negligible loss of classification accuracy (see Section~\ref{sec:exp}), the maximal compactness is limited by the filter size $L$, as mentioned in Section~\ref{sec:spatial-sampling}.

	In order to seek more compact networks without such limitation, we propose a channel sharing strategy for WSNet to learn by weight sampling along the channel dimension. As illustrated in Figure~\ref{fig:illustration} (top panel), the actual filter used in convolution is generated by  repeating sampling for $C$ times. The relation between the channels of filters before and after channel sampling is:
	\begin{equation}
	\label{activation}
	M = M^*\times C,
	\end{equation}
	Therefore, the \textit{compactness} of WSNet along the channel dimension achieves $C$.
	As introduced later in Experiments (Section~\ref{sec:exp}), we observe that the repeated weight sampling along the channel dimension significantly reduces the model size of WSNet without significant performance drop. One notable advantage of channel sharing is that the maximum compactness can be as large as $M$ (\textit{i.e.} when the condensed filter has channel of 1), which paves the way for learning much more aggressively smaller models (\textit{e.g.} more than 100$\times$ smaller models than baselines). We attribute the effectiveness of channel sharing to reducing the redundancy along the channel dimension, especially in top layers. In general architecture design, the number of filter channels grows linearly with the layer depth. However, the spatial size of kernels becomes smaller or remains unchanged. This implies redundancy in higher layers mainly come from the channel dimension.

	The above analysis for weight sampling along spatial/channel dimensions can be conveniently generalized from convolution layers to fully connected layers. For a fully connected layer, we treat its weights as a flattened vector with channel of 1, along which the spatial sampling (ref.\   Section~\ref{sec:spatial-sampling}) is performed to reduce the size of learnable parameters. For more details, please refer the supplementary material.

	\subsubsection{The training of condensed filters}
	WSNet is trained from the scratch in a similar way to conventional deep convolutional networks by using standard error back-propagation. Since every weight $\mathbf{K}_{l,m,n}$ in the convolutional kernel $\mathbf{K}$ is sampled from the condensed filter $\mathbf{\Phi}$ along the spatial and channel dimension, the only difference is the gradient of $\mathbf{\Phi}_{i,j}$ is the summation of all gradients of weights that are tied to it. Therefore, by simply recording the position mapping $\mathcal{M}: (i,j) \rightarrow (l,m,n)$ from $\mathbf{\Phi}_{i,j}$ to all the tied weights in $\mathbf{K}$, the gradient of $\mathbf{\Phi}_{i,j}$ is calculated as:
	\begin{equation}
	\label{eq:grad-phi}
	\frac{\partial \mathcal{L}}{\partial \mathbf{\Phi}_{i,j}} = \sum_{s \in \mathcal{M}(i,j)}\frac{\partial \mathcal{L}}{\partial \mathbf{K}_{s}}
	\end{equation}
	\noindent where $\mathcal{L}$ is the conventional cross-entropy loss function. In open-sourced machine learning libraries which represent computation as graphs, such as TensorFlow~\cite{tensorflow}, Equation~\eqref{eq:grad-phi} can be calculated automatically.
	\subsection{Denser Weight Sampling}
	{The performance of WSNet might be adversely affected when the size of condensed filter is decreased aggressively (\textit{i.e.} when $S$ and $C$ are large). To enhance the learning capability of WSNet, we could sample more filters from the condensed filter. Specifically, we use a smaller sampling stride $\bar{S}$ ($\bar{S} < S$) when performing spatial sampling. In order to keep the shape of weights unchanged in the following layer, we append a 1$\times$1 convolution layer with the shape of $(1, \bar{n}, n)$ to reduce the channels of densely sampled filters. It is experimentally verified that denser weight sampling can effectively improve the performance of WSNet in Section~\ref{sec:exp}. However, since it also brings extra parameters and computational cost to WSNet, denser weight sampling is only used in lower layers of WSNet whose filter number ($n$) is small. Besides, one can also conduct channel sampling on the added 1$\times$1 convolution layers to further reduce their sizes.}

	\begin{figure}
		\centering
		\includegraphics[width=\linewidth]{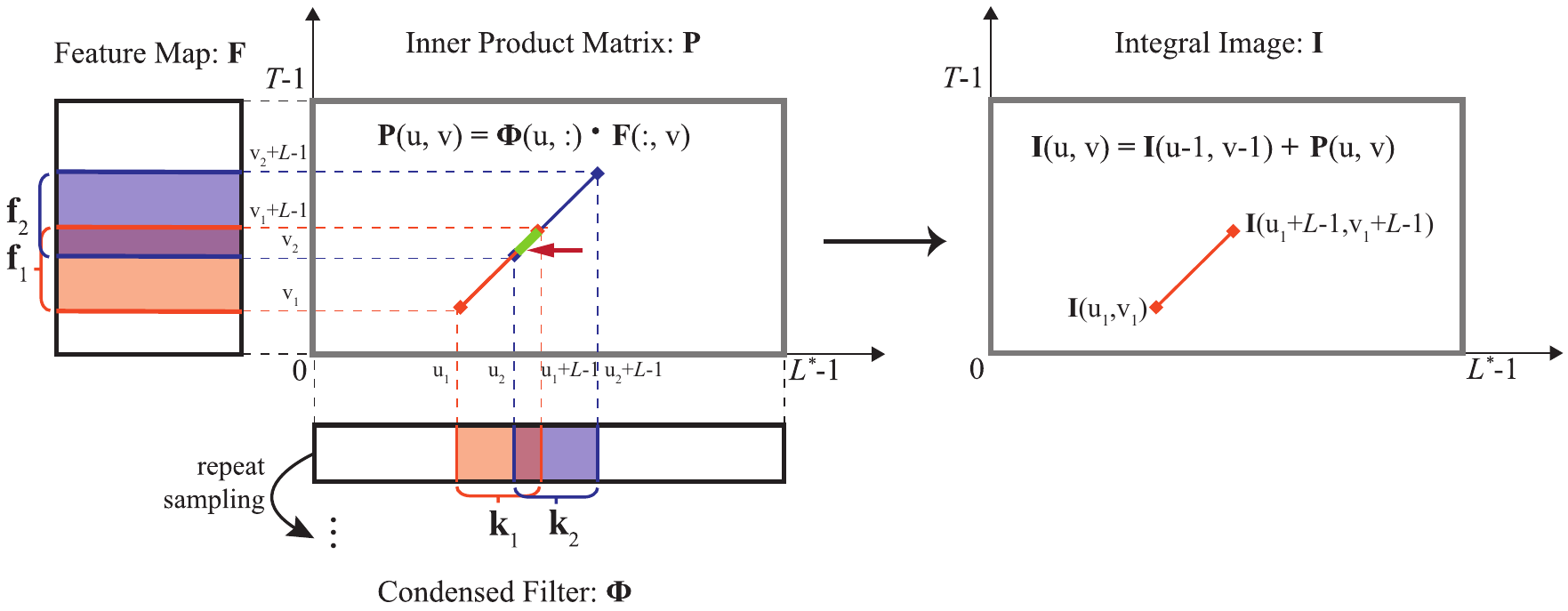}
		\caption{Illustration of efficient computation with integral image in WSNet. The inner product map $\mathbf{P} \in \mathbb{R}^{T\times L^*}$ calculates the inner product  of each row in $\mathbf{F}$ and each column in $\mathbf{\Phi}$ as in Eq.~\eqref{eq:P}. The convolution result between a filter $\mathbf{k}_1$ which is sampled from $\mathbf{\Phi}$ and the input patch $\mathbf{f}_1$ is then the summation of all values in the segment between $(u,v)$ and $(u+L-1, v+L-1)$ in $\mathbf{P}$ (recall that $L$ is the convolutional filter size). Since there are repeated calculations when the filter and input patch are overlapped, \textit{e.g.} the green segment indicated by arrow when performing convolution between $\mathbf{k}_2$ and $\mathbf{s}_2$,  we construct the integral image $\mathbf{I}$ using $\mathbf{P}$ according to Eq.~\eqref{eq:I}. Based on $\mathbf{I}$, the convolutional results between any sampled filter and input patch can be retrieved directly in time complexity of O(1) according to Eq.~\eqref{eq:new_G}, \textit{e.g.} the results of $\mathbf{k}_1 \cdot \mathbf{s}_1$ is $\mathbf{I}(u_1 + L-1, v_1 + L-1) - \mathbf{I}(u_1-1,v_1-1)$. For notation definitions, please refer to Sec.~\ref{sec:notation}. The comparisons of computation costs between WSNet and the baselines using conventional architectures are introduced in Section~\ref{sec:integral-image}. }
		\label{fig:integral-image}
	\end{figure}

	\subsection{Efficient Computation with integral image}
	\label{sec:integral-image}
	According to Equation~\ref{eq:conv}, the computation cost in terms of the number of multiplications and adds (\textit{i.e.} Mult-Adds) in a conventional convolutional layer is:
	\begin{equation}
	\label{eq:conventional_cost}
	TMLN
	\end{equation}

	However,  as illustrated in Figure~\ref{fig:integral-image}, since all filters in a layer in WSNet are sampled from a condensed filter $\mathbf{\Phi}$ with stride $S$, calculating the results of convolution in the conventional way as in Eq.~\eqref{eq:conv} incurs severe computational redundancies. Concretely, as can be seen from Eq.~\eqref{eq:conv}, one item in the ouput feature map is equal to the summation of $L$ inner products between the row vector of $\mathbf{f}$ and the column vector of $\mathbf{k}$. Therefore, when two overlapped filters that are sampled from the condensed filter (\textit{e.g.} $\mathbf{k}_1$ and $\mathbf{k}_2$ in Fig.~\ref{fig:integral-image}) convolves with the overlapped input windows (\textit{e.g.} $\mathbf{f}_1$ and $\mathbf{f}_2$ in Fig.~\ref{fig:integral-image})), some partially repeated calculations exist (\textit{e.g.} the calculations highlight in green and indicated by arrow in Fig.~\ref{fig:integral-image}). To eliminate such redundancy in convolution and speed-up WSNet, we propose a novel integral image method to enable efficient computation via sharing computations.

	We first calculate an inner product map $\mathbf{P} \in \mathbb{R}^{T\times L^*}$ which stores the inner products between each row vector in the input feature map (\textit{i.e.} $\mathbf{F}$) and each column vector in the condensed filter (\textit{i.e.} $\mathbf{\Phi}$):
	\begin{equation} \label{eq:P}
	\mathbf{P}(u,v) =
	\begin{cases}
	\mathbf{F}_{u,:} \cdot\mathbf{\Phi}_{:,v}, & u\in [0, T-1] \ \text{and}\ v\in [0, L^*-1] \\
	0, & otherwise.
	\end{cases}
	\end{equation}

	The integral image for speeding-up convolution is denoted as $\mathbf{I}$. It has the same size as $\mathbf{P}$ and can be conveniently obtained throught below formulation:
	\begin{align}\label{eq:I}
	\mathbf{I}(u,v) =
	\begin{cases}
	\mathbf{I}(u-1, v-1) + \mathbf{P}(u,v), & u > 0, v > 0\\
	\mathbf{P}(u,0), & v = 0 \\
	\mathbf{P}(0,v), & u = 0
	\end{cases}
	\end{align}

\begin{figure}
	\centering
	\includegraphics[width=0.8\linewidth]{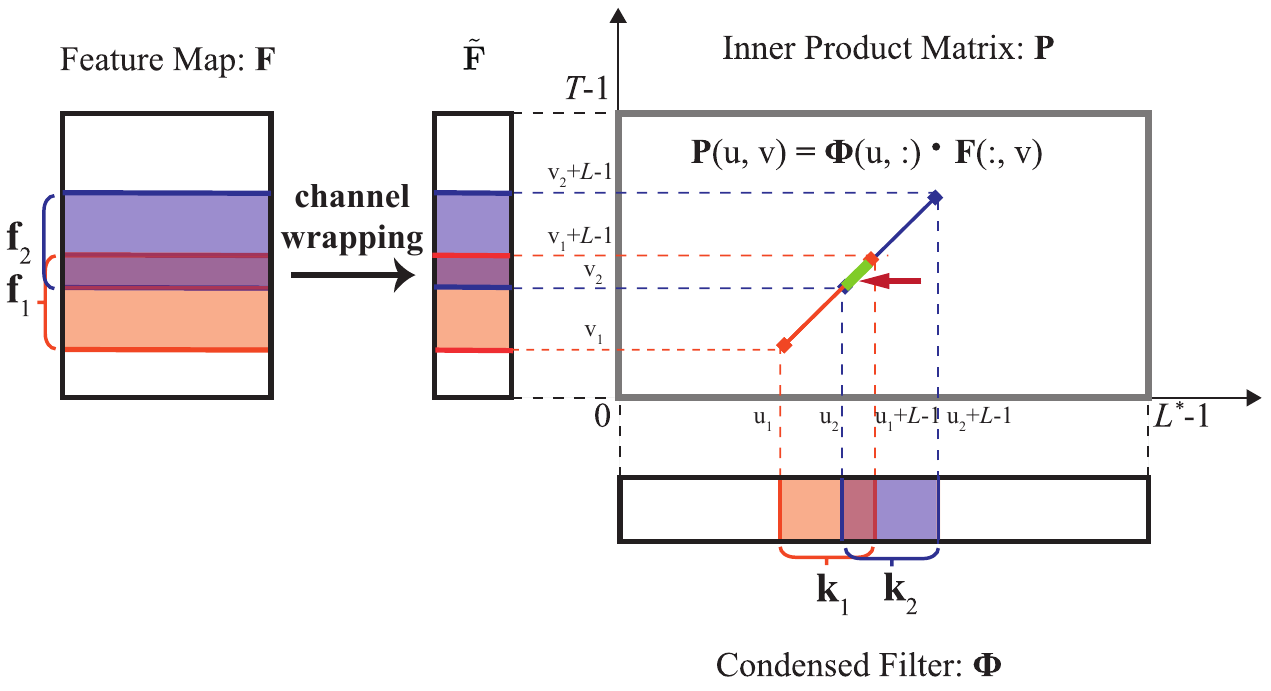}
	\caption{A variant of the integral image method used in practice which is more efficient than that illustrated in Figure~\ref{fig:integral-image}. Instead of repeatedly sampling along the channel dimension of $\mathbf{\Phi}$ to convolve with the input $\mathbf{F}$, we wrap the channels of $\mathbf{F}$ by summing up $C$ matrixes that are evenly divided from $\mathbf{F}$ along the channels, \textit{i.e.} $\mathbf{\tilde F}(i,j) = \sum_{c=0}^{C-1}\mathbf{F}(i,j+cM^*)$. Since the channle of $\tilde{\mathbf{F}}$ is only $1/C$ of the channel of $\mathbf{F}$, the overall computation cost is reduced as demonstrated in Eq.~\eqref{eq:integral_cost_2}.}
	\label{fig:integral-image-v2}
\end{figure}

	Based on $\mathbf{I}$, all convolutional results can be obtained in time complexity of $O(1)$ as follows
	\begin{equation}\label{eq:new_G}
	\mathbf{G}_{t,n} = \mathbf{I}(t + L-1, nS + L-1) - \mathbf{I}(t-1, nS-1)
	\end{equation}
	Recall that the $n$-th filter lies in the spatial range of $(nS, nS + L-1)$ in the condensed filter $\mathbf{\Phi}$. Since $\mathbf{G} \in \mathbb{R}^{T \times N}$, it thus takes $TN$ times of calculating Eq.~\eqref{eq:new_G} to get $\mathbf{G}$.
	In Eq.~\eqref{eq:P} $\sim$ Eq.~\eqref{eq:new_G}, we omit the case of padding for clear description. When zero padding is applied, we can freely get the convolutional results for the padded areas even without using Eq.~\eqref{eq:new_G} since $\mathbf{I}(u,v) = \mathbf{I}(T,v-1), u>T$.

	Based on Eq.~\eqref{eq:P} $\sim$ Eq.~\eqref{eq:new_G}, the computation cost of the proposed integral image method is
	\begin{equation} \label{eq:integral_cost}
	\underbrace{TML^*}_{\text{Eq.~\eqref{eq:P}}}+\underbrace{TL^*}_{\text{Eq.~\eqref{eq:I}}}+\underbrace{TN}_{\text{Eq.~\eqref{eq:new_G}}} = T(M+1)L^* + TN.
	\end{equation}

	Note the computation cost of $\mathbf{P}$ (\textit{i.e.} Eq.~\eqref{eq:P}) is the dominating term in Eq.~\eqref{eq:integral_cost}. Based on Eq.~\eqref{eq:conventional_cost}, Eq.~\eqref{eq:integral_cost} and Eq.~\eqref{eq:s_comp}, the theoretical acceleration ratio is
	\begin{equation*}
	\frac{TMLN}{T(M+1)L^* + TN} \approx \frac{L}{S}
	\end{equation*}

	Recall that $L$ is the filter size and $S$ is the pre-defined stride when sampling filters from the condensed filter $\mathbf{\Phi}$ (ref. to Eq.~\eqref{eq:s_comp}).

	In practice, we adopt a variant of the above method to further boost the computation efficiency of WSNet, as illustrated in Fig~\ref{fig:integral-image-v2}. In Eq.~\eqref{eq:P}, we repeat $\mathbf{\Phi}$ by $C$ times along the channel dimension to make it equal with the channel of the input $\mathbf{F}$. However, we could first wrap the channels of $\mathbf{F}$ by accumulating the values with interval of $L$ along its channel dimension to a thinner feature map $\mathbf{\tilde F} \in \mathbb{R}^{T\times M^*}$ which has the same channel number as $\mathbf{\Phi}$, \textit{i.e.} $\mathbf{\tilde F}(i,j) = \sum_{c=0}^{C-1}\mathbf{F}(i,j+cM^*)$. Both Eq.~\eqref{eq:I} and Eq.~\eqref{eq:new_G} remain the same. Then the computational cost is reduced to
	\begin{equation} \label{eq:integral_cost_2}
	\underbrace{TM^*(C-1)}_{\text{channel warp}} + \underbrace{TM^*L^*}_{\text{Eq.~\eqref{eq:P}}}+\underbrace{TL^*}_{\text{Eq.~\eqref{eq:I}}}+\underbrace{TN}_{\text{Eq.~\eqref{eq:new_G}}}
	\end{equation}
	where the first item is the computational cost of warping the channels of $\mathbf{F}$ to obtain $\mathbf{\tilde{F}}$. Since the dominating term (\textit{i.e.} Eq.~\eqref{eq:P}) in Eq~\eqref{eq:integral_cost_2} is smaller than in Eq.~\eqref{eq:integral_cost}, the overall computation cost is thus largely reduced. By combining Eq.~\eqref{eq:integral_cost_2} and Eq.~\eqref{eq:conventional_cost}, the theoretical acceleration compared to the baseline is
	\begin{equation}
	\frac{MLN}{M^*(C+ L^* -1)+(L^*+N)}
	\end{equation}

	Finally, we note that the integral image method applied in WSNet naturally takes advantage of the property in weight sampling: redundant computations exist between overlapped filters and input patches. Different from other deep model speedup methods~\citep{sindhwani2015structured,denton2014exploiting} which require to solve time-consuming optimization problems and incur performance drop, the integral image method can be seamlessly embeded in WSNet without negatively affecting the final performance.

\subsection{An Intuitive Extension of WSNet from 1D convnet to 2D convnet}
In this paper, we focus on WSNet with 1D convnets. Comprehensive experiments clearly demonstrate its advantages in learning compact and computation-efficient networks. We note that WSNet is general and can also be applied to build 2D convnets. In 2D convnets, each filter has three dimensions including two spatial dimensions (\textit{i.e.} along X and Y directions) and one channel dimension. One straightforward extension of WSNet to 2D convnets is as follows: for spatial sampling, each filter is sampled out as a patch (with the same number of channels as in condensed filter) from condensed filter. Channel sampling remains the same as in 1D convnets, \textit{i.e.} repeat sampling in the channel dimension of condensed filter. Following the notations for WSNet with 1D convnets (ref. to Sec.~\ref{sec:notation}), we denote the filters in one layer as $\mathbf{K} \in \mathbb{R}^{w\times h\times M \times N}$ where $(w, h, M, N)$ denote the width and height of each filter, the number of channels and the number of filters respectively. The condensed filter $\mathbf{\Phi}$ has the shape of $(W, H, M^*)$. The relations between the shape of condensed filter and each sampled filter are:
\begin{equation}\label{eq:2D-spatial}
\begin{aligned}
W &= w + (\lceil\sqrt{N}\rceil  - 1)S_w \\
H &= h + (\lceil\sqrt{N}\rceil - 1) S_h \\
M &= M^* \times C
\end{aligned}
\end{equation}
\noindent where $S_w$ and $S_h$ are the sampling strides along two spatial dimensions and $C$ is the compactness of WSNet along channel dimension. The compactnesses (ref. to Eq.~\eqref{eq:compactness} for denifinition) of WSNet along spatial and channel dimension are $\frac{WH}{whN}$ and $C$ respectively. However, such straightforward extension of WSNet to 2D convnets may not be optimum and we believe there are more sophisticated and effective methods for applying WSNet to 2D convnets and we would like to explore in our future work. Nevertheless, we conduct preliminary experiments on 2D convents using above intuitive extension and verify the effectiveness of WSNet in image classification tasks (on MNIST and CIFAR10).

	\begin{table*}
		\caption{Baseline-1:  configurations of the baseline network used on MusicDet200K. Each convolutional layer is followed by a nonlinearity layer (\textit{i.e.} ReLU), batch normalization layer and pooling layer,  which are omitted in the table for brevity. The strides of all pooling layers are 2. The padding strategies adopted for both convolutional layers and fully connected layers are all ``size preserving".}
		\label{table:cnn-1}
		\centering
		\footnotesize
		\begin{tabular}{L{2.4cm}|C{0.85cm}|C{0.85cm}|C{0.85cm}|C{0.85cm}|C{0.85cm}|C{0.85cm}|C{0.85cm}|c|c} \toprule
			Layer & conv1 & conv2 & conv3 & conv4 & conv5 & conv6 & conv7 & fc1 &  fc2 \\
			\midrule
			Filter sizes   & 32 & 32 & 16   & 8     & 8     & 8    & 4    & 1536 & 256 \\
			\#Filters      & 32 & 64 & 128  & 128   & 256   & 512  & 512  & 256  & 128  \\
			Stride         & 2  & 2  & 2    & 2     & 2     & 2    & 2    & 1    & 1    \\
			\#Params    & 1K & 65K & 130K & 130K & 260K & 1M & 1M & 390K & 33K \\
			\#Mult-Adds ($10^8$) & 4.1 & 65.5 & 32.7 & 8.2 & 4.1 & 4.2 & 1.0 & 0.1 & 0.007 \\
			\bottomrule
		\end{tabular}
	\end{table*}

	\begin{table*}
		\caption{Baseline-2:  configuration of the baseline network used on ESC-50, UrbanSound8K and DCASE. This baseline is adapted from SoundNet~\citep{soundnet} by applying pooling layers to all but the last convolutional layer. For brevity, the nonlinearity layer (\textit{i.e.} ReLU), batch normalization layer and pooling layer following each convolutional layer are omitted. The kernel sizes for pooling layers following conv1-conv4 and conv5-conv7 are 8 and 4 respectively. The stride of every pooling layers is 2.}
		\label{table:cnn-2}
		\centering
		\footnotesize
		\begin{tabular}{L{3.2cm}|c|c|c|c|c|c|c|c} \toprule
			Layer & conv1 & conv2 & conv3 & conv4 & conv5 & conv6 & conv7 & conv8\\
			\midrule
			Filter sizes  & 64 & 32 & 16   & 8     & 4     & 4    & 4     & 8 \\
			\#Filters      & 16 & 32 & 64   & 128 & 256 & 512 & 1024 & 1401 \\
			Stride          & 2  & 2   & 2     & 2     & 2     & 2    & 2    & 2     \\
			\#Params    & 1K & 16K & 32K & 65K & 130K & 520K & 2M & 11M \\
			\#Mult-Adds ($10^8$) & 2.3 & 9.0 & 4.5 & 2.3 & 1.2 & 1.2 & 1.2 & 2.3 \\
			\bottomrule
		\end{tabular}
	\end{table*}

		\begin{table*}[!htbp]
		\caption{Ablative study of the effects of different settings of WSNet on the model size, computation cost (in terms of \#mult-adds) and classification accuracy on ESC-50. For clear description, we name WSNets with different settings by the combination of symbols S/C/D/Q. ``S'' denotes the weight sampling along spatial dimension; ``C'' denotes the weight sampling along the channel dimension. ``D'' denotes denser filter sampling. ``Q'' denotes weight quantization. The numbers in subscripts of S/C/D/Q denotes the maximum compactness (ref. to Sec.~\ref{sec:notation} for the definition of compactness) on spatial/channel dimension in all layers, the ratio of the number of filters in WSNet versus in the baseline and the ratio of WSNet's size before and after weight quantization, respectively. The model size and the computational cost are provided for the baseline. For the model size and \#mult-adds of WSNet, we provide the ratio of the baseline's model size versus WSNet's model size and the ratio of the baseline's \#Mult-Adds versus WSNet's \#Mult-Adds.}
		\label{table:ablation}
		\centering
		\scriptsize
		\begin{tabular}{p{1.3cm}|p{0.05cm}p{0.05cm}p{0.1cm}|p{0.05cm}p{0.05cm}p{0.1cm}|p{0.05cm}p{0.05cm}p{0.1cm}|p{0.05cm}p{0.05cm}p{0.1cm}|p{0.05cm}p{0.05cm}p{0.1cm}|C{1.2cm}|C{1.2cm}|c} \toprule
			\multirow{2}{*}{WSNet's}   & \multicolumn{3}{c@{}|}{conv\{1-4\}}  & \multicolumn{3}{c@{}|}{conv5}  & \multicolumn{3}{c@{}|}{conv6} & \multicolumn{3}{c@{}|}{conv7} & \multicolumn{3}{c@{}|}{conv8} & \multirow{3}{*}{Acc.}& \multirow{2}{*}{Model} & \multirow{3}{*}{Mult-Adds}\\
			\cmidrule(lr){2-4} \cmidrule(lr){5-7} \cmidrule(lr){8-10} \cmidrule(lr){11-13} \cmidrule(lr){14-16}
			settings             &S&C&D            &S&C&D      &S&C&D      &S&C&D      &S&C&D      &  & size& \\
			\midrule
			Baseline                  & 1  & 1  & 1    & 1 & 1 & 1    & 1 & 1 & 1    & 1 & 1 & 1   & 1 & 1 & 1    & 66.0 $\pm$ 0.2&  13M (1$\times$) & 2.4e8 (1$\times$)\\
			BaselineQ$_4$                 & 1  & 1  & 1    & 1 & 1 & 1    & 1 & 1 & 1    & 1 & 1 & 1   & 1 & 1 & 1    & 65.7 $\pm$ 0.2& 4$\times$ & 1$\times$\\
			\midrule
			S$_2$                              & 2  & 1  & 1    & 2 & 1 & 1    & 2 & 1 & 1    & 2 & 1 & 1   & 2 & 1 & 1    & 66.6 $\pm$ 0.3  &  2$\times$ & 1$\times$\\
			S$_4$                              & 4  & 1  & 1    & 4 & 1 & 1    & 4 & 1 & 1    & 4 & 1 & 1   & 4 & 1 & 1    & 66.3 $\pm$ 0.1  &  4$\times$ & 1.6$\times$\\
			S$_8$                             & 8  & 1  & 1    & \underline{4} & 1 & 1    & \underline{4} & 1 & 1    & \underline{4} & 1 & 1    & 8 & 1 & 1      & 65.2 $\pm$ 0.1  &  7$\times$ & 4.7$\times$\\
			\midrule
			C$_2$                              & 1  & 2  & 1    & 1 & 2 & 1    & 1 & 2 & 1    & 1 & 2 & 1   & 1 & 2 & 1    & 66.8 $\pm$ 0.2  &  2$\times$ & 1$\times$\\
			C$_4$                              & 1  & 4  & 1    & 1 & 4 & 1    & 1 & 4 & 1    & 1 & 4 & 1   & 1 & 4 & 1    & 66.5 $\pm$ 0.3  &  4$\times$ & 1.6$\times$\\
			C$_8$                              & 1  & 8  & 1    & 1 & 4 & 1    & 1 & 4 & 1    & 1 & 4 & 1    & 1 & 8 & 1   & 65.8 $\pm$ 0.3  &  8$\times$ & 2.8$\times$\\
			\midrule
			S$_4$C$_4$ & 4  & 4  & 1    & 4 & 4 & 1    & 4 & 4 & 1    & 4 & 4 & 1   & 4 & 4 & 1    & 65.6 $\pm$ 0.3  &  16$\times$ &6.3$\times$\\
			S$_8$C$_8$ & 4  & 4 & 1    & 4  & 8 & 1    & 4  & 8 & 1   & 4  & 8 & 1   & 8  & 8 & 1   & 65.2 $\pm$ 0.3  &  60$\times$ & \textbf{18.1$\times$}\\
			\midrule
			S$_8$C$_4$D$_2$                 & 4  & 4  & 2    & 4 & 4 & 1    & 4 & 4 & 1    & 4 & 4 & 1   & 8 & 4 & 1 & \textbf{66.5 $\pm$ 0.1}  &  25$\times$ & 2.3$\times$\\
			S$_8$C$_4$D$_2$Q$_4$             & 4  & 4  & 2    & 4 & 4 & 1    & 4 & 4 & 1    & 4 & 4 & 1   & 8 & 4 & 1 & 66.2 $\pm$ 0.1  &  100$\times$ & 2.3$\times$\\
			S$_8$C$_8$D$_2$ & 4  & 4 & 1    & 4  & 8 & 1    & 4  & 8 & 1   & 4 & 8 & 1   & 8  & 8 & 1   & 66.1 $\pm$ 0.0  &  45$\times$ & \textbf{2.4$\times$}\\
			S$_8$C$_8$D$_2$Q$_4$   & 4  & 4  & 2    & 4 & 8 & 1    & 4 & 8 & 1    & 4 & 8 & 1   & 8 & 8 & 1    &65.8 $\pm$ 0.0  &  \textbf{180$\times$} & 2.4$\times$\\
			\bottomrule
		\end{tabular}
	\end{table*}

	\section{Experiments} \label{sec:exp}

	\subsection{Experimental Settings} \label{sec:setting}
	\paragraph{Datasets and baseline networks}
	We collect a large-scale music detection dataset (MusicDet200K) from publicly available platforms (\textit{e.g.} Facebook, Twitter, \textit{etc.}) for conducting experiments. For fair comparison with previous literatures, we also test WSNet on three standard, publicly available datasets, \textit{i.e} ESC-50, UrbanSound8K and DCASE. Due to space limit, please refer to the details of used datasets in supplementary material.

	To test the scability of WSNet to different network architectures (\textit{e.g.} whether having fully connected layers or not), two baseline networks are used in comparision. Their architectures are shown in Table~\ref{table:cnn-1} and Table~\ref{table:cnn-2} respectively.
	\paragraph{Evaluation criteria} To demonstrate that WSNet is capable of learning more compact and efficient models than conventional CNNs, three evaluation criteria are used in our experiments: model size, the number of multiply and adds in calculation (mult-adds) and classification accuracy. For the results of WSNet models, we also give the std of five different runs.

	\paragraph{Implementation details} WSNet is implemented and trained from scratch in Tensorflow~\citep{tensorflow}.  Following~\cite{soundnet}, the Adam~\citep{adam} optimizer, a fixed learning rate of
	0.001, and momentum term of 0.9 and batch size of 64 are used throughout experiments. We initialized all the weights to zero mean gaussian
	noise with a standard deviation of 0.01. In the network used on MusicDet200K, the dropout ratio for the dropout layers~\citep{dropout} after each fully connected layer is set to be 0.8. The overall training takes 100,000 iterations.

	\subsection{Results and analysis}
	\label{sec:results}
	\subsubsection{ESC-50} \label{sec:}
	\paragraph{Ablation analysis} We investigate the effects of each component in WSNet on the model size, computational cost and classification accuracy. The comparative study results of different settings of WSNet are listed in Table~\ref{table:ablation}. For clear description, we name WSNets with different settings by the combination of symbols S/C/D/Q. Please refer to the caption of Table~\ref{table:ablation} for detailed meanings.

	\qquad \textit{(1) Spatial sampling.} We test the performance of WSNet by using different sampling stride $S$ in spatial sampling. As listed in Table~\ref{table:ablation}, S$_2$ and S$_4$ slightly outperforms the classification accuracy of the baseline, possibly due to reducing the overfitting of models. When the sampling stride is 8, \textit{i.e.} the compactness in spatial dimension is 8 (ref. to Section~\ref{sec:spatial-sampling}), the classification accuracy of S$_8$ only drops by 0.6\%. Note that the maximum compactness along the spatial dimension is equal to the filter size, thus for the layer ``conv\{5-7\}'' which have filter sizes of 4, their compactnesses are limited by 4 (highlighted by underlines in Table~\ref{table:ablation}). Above results clearly demonstrate that the spatial sampling enables WSNet to learn significantly smaller model with comparable accuracies w.r.t. the baseline.

	\qquad \textit{(2) Channel sampling.} Three different compactness along the channel dimension, \textit{i.e.} 2, 4 and 8 are tested by comparing with baslines. It can be observed from Table~\ref{table:ablation} that C$_2$ and C$_4$ and C$_8$ have linearly reduced model size without incurring noticeable drop of accuracy. In fact, C$_2$ and C$_4$ can even improve the accuracy upon baselines, demonstrating the effectiveness of channel sampling in WSNet. {When learning more compact models, C$_8$ demonstrates better performance compared to S$_8$ that has the same compactness in the spatial dimension, which suggests we should focus on the channel sampling when the compactness along the spatial dimension is high.}

	We then simultaneously perform weight sampling on both the spatial and channel dimensions. As demonstrated by the results of S$_4$C$_4$
	and S$_8$C$_8$, WSNet can learn highly compact models without significant performance drop (less than 1\%).

	\qquad \textit{(3) Denser weight sampling.}  Denser weight sampling is used to enhance the learning capability of WSNet with aggressive compactness (\textit{i.e.} when $S$ and $C$ are large) and make up the performance loss caused by sharing too much parameters among filters. As shown in Table~\ref{table:ablation}, by sampling 2$\times$ more filters in conv\{1-4\}, S$_8$C$_8$D$_2$ significantly outperforms the S$_8$C$_8$. Above results demonstrate the effectiveness of denser weight sampling to boost the performance.

	\qquad \textit{(4) Integral image for efficient computation.} As evidenced in the last column in Table~\ref{table:ablation}, the proposed integral image method consistently reduces the computation cost of WSNet. For S$_8$C$_8$ which is 60$\times$ smaller than the baseline, the computation cost (in terms of \#mult-adds) is significantly reduced by 18.1 times. Due to the extra computation cost brought by the 1$\times$1 convolution in denser filter sampling, S$_8$C$_8$D$_2$ achieves lower acceleration (2.4$\times$). Group convolution~\citep{resnext} can be used to alleviate the computation cost of the added 1$\times$1 convolution layers. We will explore this direction in our future work.

	\qquad \textit{(5) Weight quantization.}
It can be observed from Table~\ref{table:ablation} that by using 256 bins to represent each weight by one byte (\textit{i.e.} 8bits), S$_8$C$_8$D$_2$Q$_4$ and S$_8$C$_4$D$_2$Q$_4$ have much smaller model size compared with baselines while incurring negligible accuracy loss. The above result demonstrates that the weight quantization is complementary to WSNet and they can be used jointly to effectively reduce the model size of WSNet. Please ref. to supplementary material for the details of the weight quantization methods.

\begin{figure}\label{fig:narrow-baseline}
			\centering
			\includegraphics[width=0.8\linewidth]{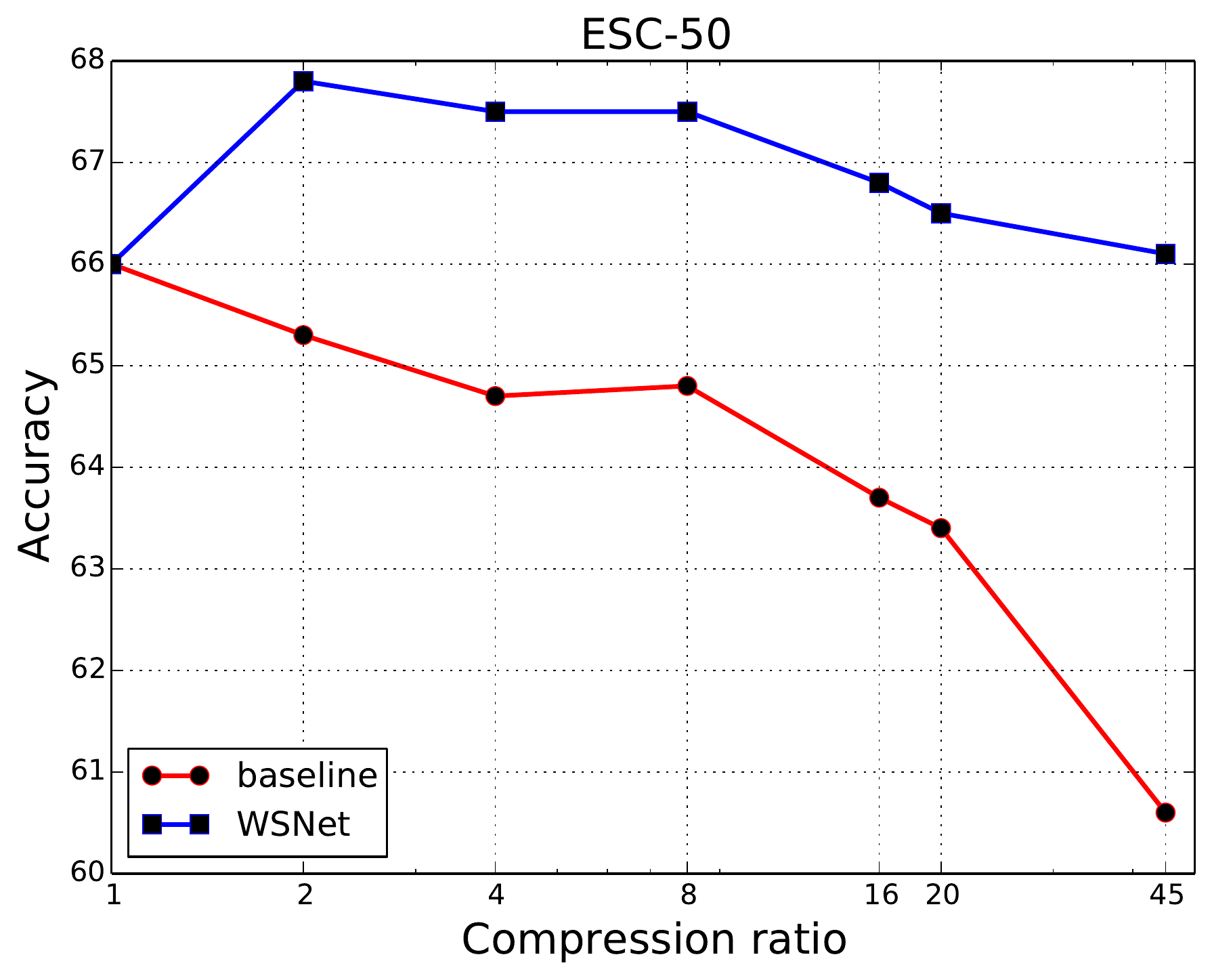}
			\caption{The accuracies of baselines and WSNet with the same model size on ESC-50 dataset. Note the compression ratios (or compactness for WSNet) are shown in log scale. }
			\label{fig:acc-comp1}
	\end{figure}

	\qquad \textit{(6) WSNet versus narrowed baselines.} 	To further verify WSNet's capacity of learning compact models, we compare WSNet with baselines compressed in an intuitive way, \textit{i.e.} reducing the number of filters in each layer. If \#filters in each layer is reduced by $T$, the overall \#parameters in baselines is reduced by $T^2$ (\textit{i.e.} the compression ratio of model size is $T^2$). In Figure~\ref{fig:acc-comp1}, we plot how baseline accuracy varies with respect to different compression ratios and the accuracies of WSNet with the same model size of compressed baselines.

	As shown in Figure~\ref{fig:acc-comp1}, WSNet outperforms baselines by a large margin across all compression ratios. Particularly, when the compression ratios are large (\textit{e.g.} 45), baselines suffer severe performance drop. In contrast, WSNet achieves comparable accuracies with full-size baselines (66.1 versus 66.0). This clearly demonstrates the effectiveness of weight sampling methods proposed in WSNet. In supplementary material, we also present the comparison between WSNet and narrowed baselines on MusicDet200K.

	\subsubsection{Comparison with State-of-the-art}
	The comparison of WSNet with other state-of-the-arts on ESC-50 is listed in Table~\ref{table:esc50}. Compared with the SoundNet trained with provided data, WSNets significantly outperform its classification accuracy by over 10\% with more than 100$\times$ smaller models. After pre-training using a large number of unlabeled videos, SoundNet$^*$ achieves better accuracy than WSNet. However, since the unsupervised pre-training method is orthogonal to WSNet, we believe that WSNet can achieve better performance by training in a similar way as SoundNet~\citep{soundnet} on a large amount of unlabeled video data. Due to space limit, for experimental results on other datasets as well as the ablative study on MusicDet200K, please refer to supplementary material.

	\subsection{Discussions} \label{sec:motivation}
	We argue that there are two reasons for the success of WSNet:
	\textbf{(1)} The epitome methods~\cite{benoit2011sparse,aharon2008sparse,jojic2003epitomic} have been successfully deployed in sparse coding literatures, where the coding dictionaries are formed by overlapping patches in the epitome which has few free parameters. This indicates effective representations of complex signals can be generated from a low-dimensional space (with high parameter efficiency). It thus motivates us to learn compact (or epitomic) filters in deep neural networks, \textit{i.e.} all filters which participate in the actual convolution are generated from the condensed filters.
	\textbf{(2)} Weight quantization techniques were successfully applied for compressing deep models where multiple weights are encoded into the same value. WSNet goes further to overcome limitations of existing quantization methods through capturing the common correlations among learned filters. For example, filters of the first layer in SoundNet~\cite{soundnet} (as illustrated in Figure 5 in \cite{soundnet}) all learn similar constituent patterns, \textit{e.g.} the descending/ascending slope lines. The proposed weight sampling method enables WSNet to learn shared patterns by explicitly sampling filters from the condensed filter with overlapping. At the same time, the non-overlapped parts of sampled filters are able to learn different features which endows WSNet with strong learning capabilities. This is the \textbf{main reason} that why WSNet can learn much smaller networks without noticeable performance drop compared to baselines. Moreover, as the sampled filters are overlapped, we could use an integral image based method to speed up WSNets (ref. to Section~\ref{sec:integral-image}). In this way, WSNet is able to learn both smaller and faster networks effectively.

	\begin{table}
	\caption{Comparison with state-of-the-arts using 1D CNNs on ESC-50. All results of WSNet are obtained by 10-folder validation. Please refer to Table~\ref{table:ablation} for the meaning of symbols S/C/D/Q. SoundNet$^*$ use extra training data while other methods use only provided training data.}
	\label{table:esc50}
	\footnotesize
	\begin{tabular}{L{4.1cm}ccc} \toprule
		Model  & Acc. (\%) & Model size\\
          \midrule
          Piczak ConvNet~\citep{piczak2015environmental} & 64.5 & 28M\\
          SoundNet~\citep{soundnet} & 51.1 & 13M \\
          SoundNet$^*$~\citep{soundnet} & 72.9 & 13M \\
          \midrule
          WSNet (S$_8$C$_4$D$_2$) & 66.5 $\pm$ 0.10 & 0.52M \\
          WSNet (S$_8$C$_4$D$_2$Q$_4$) & \textbf{66.25 $\pm$ 0.25} & \textbf{0.13M} \\
          WSNet (S$_8$C$_8$D$_2$) & 66.1 $\pm$ 0.15 & 0.29M\\
          WSNet (S$_8$C$_8$D$_2$Q$_4$) & \textbf{65.8 $\pm$ 0.25}& \textbf{0.07M} \\
          \bottomrule
	\end{tabular}
\end{table}

\begin{table}
	\caption{Test error rates (in \%) of WSNet and HashNet on CIFAR10 and MNIST. The baselines used for MNIST/CIFAR10 are simple 3-layer fully connected network and 5-layer convolutional network respectively. The model size is provided for the baseline. For the model size of WSNet/HashNet, we provide the ratio (\textit{i.e.} $n$) of the baseline's model size versus the model size of WSNet/HashNet. For WSNet, we set the layer-wise compactness to be $n$. Specifically, for each convolutional layer in WSNet, we set its compactness along spatial/channel dimension to be $\sqrt{n}/\sqrt{n}$, respectively. }
	\label{table:cifar}
	\footnotesize
	\centering
	\begin{tabular}{l|cc|cc} \toprule Model       & Model size   & Error rate  & Model size   & Error rate \\
		\midrule
		& \multicolumn{2}{c|}{CIFAR10}  & \multicolumn{2}{c}{MNIST} \\
		\midrule
		baseline            & 1.2M ($\times$)            & 14.91  & 800K (1$\times$)            & 1.37 \\
		\midrule
		HashNet		  & 16$\times$           & 21.42  & 8$\times$           & 1.43\\
		HashNet			  & 64$\times$           & 30.79 & 64$\times$           & 2.41\\
		\midrule
		WSNet			  & 16$\times$            & \textbf{17.82}  & 8$\times$            & \textbf{1.29} \\
		WSNet             & 64$\times$          &  \textbf{23.59} & 64$\times$          & \textbf{1.97}\\
		\bottomrule
	\end{tabular}
\end{table}

\subsection{Experimental results of WSNet on 2D CNNs}
Since both WSNet and HashNet~\cite{hashnet,freshnet} explore weights tying, we compare them on MNIST and CIFAR10. For fair comparison, we use the same baselines used in~\cite{hashnet,freshnet}. All hyperparameters during training follow~\cite{hashnet,freshnet}. For each dataset, we hold out 20\% of training samples to form a validation set. The comparison results between WSNet and HashNet on MNIST/CIFAR10 are listed in Table~\ref{table:cifar}, from which one can observe that when learning networks with the same sizes, WSNet achieves significantly lower error rates than HashNet on both datasets. Above results clearly demonstrate the advantages of WSNet in learning compact models.

Furthermore, we also conduct experiment on CIFAR10 with the state-of-the-art ResNet50~\cite{residual} as baseline. ResNet50 achieves top-1 accuracy of 93.03\% with \#params of 0.85M. For WSNet, we set $S_w=S_h=2$ and $C=4$. The experimental settings follow those in~\cite{residual}. WSNet is able to achieve 9$\times$ smaller model size with slight performance drop (\textbf{0.5\%}). Such promising results further demonstrate the effectiveness of WSNet.

\section{Conclusion}
In this paper, we present a class of \textbf{W}eight \textbf{S}ampling networks (WSNet) which are highly compact and efficient. A novel weight sampling method is proposed to sample filters from condensed filters which are much smaller than the independently trained filters in conventional networks. The weight sampling in conducted in two dimensions of the condensed filters, \textit{i.e.} by spatial sampling and channel sampling. Taking advantage of the overlapping property of the filters in WSNet, we propose an integral image method for efficient computation. Extensive experiments on four audio classification datasets including MusicDet200K, ESC-50, UrbanSound8K and DCASE clearly demonstrate that WSNet can learn compact and efficient networks with competitive performance.

\small
\bibliography{wsnet}

\begin{thebibliography}{51}
\providecommand{\natexlab}[1]{#1}
\providecommand{\url}[1]{\texttt{#1}}
\expandafter\ifx\csname urlstyle\endcsname\relax
  \providecommand{\doi}[1]{doi: #1}\else
  \providecommand{\doi}{doi: \begingroup \urlstyle{rm}\Url}\fi

\bibitem[Abadi et~al.(2016)Abadi, Agarwal, Barham, Brevdo, Chen, Citro,
  Corrado, Davis, Dean, Devin, et~al.]{tensorflow}
Abadi, Mart{\'\i}n, Agarwal, Ashish, Barham, Paul, Brevdo, Eugene, Chen,
  Zhifeng, Citro, Craig, Corrado, Greg~S, Davis, Andy, Dean, Jeffrey, Devin,
  Matthieu, et~al.
\newblock Tensorflow: Large-scale machine learning on heterogeneous distributed
  systems.
\newblock \emph{arXiv preprint arXiv:1603.04467}, 2016.

\bibitem[Aharon \& Elad(2008)Aharon and Elad]{aharon2008sparse}
Aharon, Michal and Elad, Michael.
\newblock Sparse and redundant modeling of image content using an
  image-signature-dictionary.
\newblock \emph{SIAM Journal on Imaging Sciences}, 1\penalty0 (3):\penalty0
  228--247, 2008.

\bibitem[Anwar et~al.(2017)Anwar, Hwang, and Sung]{anwar2015structured}
Anwar, Sajid, Hwang, Kyuyeon, and Sung, Wonyong.
\newblock Structured pruning of deep convolutional neural networks.
\newblock \emph{J. Emerg. Technol. Comput. Syst.}, 13\penalty0 (3):\penalty0
  32:1--32:18, February 2017.
\newblock ISSN 1550-4832.
\newblock \doi{10.1145/3005348}.
\newblock URL \url{http://doi.acm.org/10.1145/3005348}.

\bibitem[Aytar et~al.(2016)Aytar, Vondrick, and Torralba]{soundnet}
Aytar, Yusuf, Vondrick, Carl, and Torralba, Antonio.
\newblock Soundnet: Learning sound representations from unlabeled video.
\newblock In \emph{NIPS}, 2016.

\bibitem[Ba \& Caruana(2014)Ba and Caruana]{st2}
Ba, Jimmy and Caruana, Rich.
\newblock Do deep nets really need to be deep?
\newblock In \emph{NIPS}, 2014.

\bibitem[Bagherinezhad et~al.(2016)Bagherinezhad, Rastegari, and
  Farhadi]{bagherinezhad2016lcnn}
Bagherinezhad, Hessam, Rastegari, Mohammad, and Farhadi, Ali.
\newblock Lcnn: Lookup-based convolutional neural network.
\newblock \emph{arXiv preprint arXiv:1611.06473}, 2016.

\bibitem[Barchiesi et~al.(2015)Barchiesi, Giannoulis, Stowell, and
  Plumbley]{Barchiesi15}
Barchiesi, Daniele, Giannoulis, Diåmitrios, Stowell, Dan, and Plumbley,
  Mark~D.
\newblock Acoustic scene classification: Classifying environments from the
  sounds they produce.
\newblock \emph{IEEE Signal Processing Magazine}, 32\penalty0 (3):\penalty0
  16--34, 2015.

\bibitem[Beno{\^\i}t et~al.()Beno{\^\i}t, Mairal, Bach, and
  Ponce]{benoit2011sparse}
Beno{\^\i}t, Louise, Mairal, Julien, Bach, Francis, and Ponce, Jean.
\newblock Sparse image representation with epitomes.
\newblock In \emph{CVPR}.

\bibitem[Buciluǎ et~al.(2006)Buciluǎ, Caruana, and Niculescu-Mizil]{st1}
Buciluǎ, Cristian, Caruana, Rich, and Niculescu-Mizil, Alexandru.
\newblock Model compression.
\newblock In \emph{KDD}, 2006.

\bibitem[Cai et~al.(2006)Cai, Lu, Hanjalic, Zhang, and Cai]{Cai06}
Cai, Rui, Lu, Lie, Hanjalic, Alan, Zhang, Hong-Jiang, and Cai, Lian-Hong.
\newblock A flexible framework for key audio effects detection and auditory
  context inference.
\newblock \emph{IEEE Transactions on audio, speech, and language processing},
  14\penalty0 (3):\penalty0 1026--1039, 2006.

\bibitem[Chen et~al.(2015)Chen, Wilson, Tyree, Weinberger, and Chen]{hashnet}
Chen, Wenlin, Wilson, James, Tyree, Stephen, Weinberger, Kilian, and Chen,
  Yixin.
\newblock Compressing neural networks with the hashing trick.
\newblock In \emph{ICML}, 2015.

\bibitem[Chen et~al.(2016)Chen, Wilson, Tyree, Weinberger, and Chen]{freshnet}
Chen, Wenlin, Wilson, James~T, Tyree, Stephen, Weinberger, Kilian~Q, and Chen,
  Yixin.
\newblock Compressing convolutional neural networks in the frequency domain.
\newblock In \emph{KDD}, 2016.

\bibitem[Chen et~al.(2017)Chen, Li, Xiao, Jin, Yan, and Feng]{dpn}
Chen, Yunpeng, Li, Jianan, Xiao, Huaxin, Jin, Xiaojie, Yan, Shuicheng, and
  Feng, Jiashi.
\newblock Dual path networks.
\newblock \emph{arXiv preprint arXiv:1707.01629}, 2017.

\bibitem[Chollet(2016)]{xception}
Chollet, Fran{\c{c}}ois.
\newblock Xception: Deep learning with depthwise separable convolutions.
\newblock \emph{arXiv preprint arXiv:1610.02357}, 2016.

\bibitem[Chu et~al.(2006)Chu, Narayanan, Kuo, and Mataric]{Chu06}
Chu, Selina, Narayanan, Shrikanth, Kuo, C-C~Jay, and Mataric, Maja~J.
\newblock Where am i? scene recognition for mobile robots using audio features.
\newblock In \emph{ICME}, 2006.

\bibitem[Collins \& Kohli(2014)Collins and Kohli]{collins2014memory}
Collins, Maxwell~D and Kohli, Pushmeet.
\newblock Memory bounded deep convolutional networks.
\newblock \emph{arXiv preprint arXiv:1412.1442}, 2014.

\bibitem[Davis \& Mermelstein(1980)Davis and Mermelstein]{Davis80}
Davis, Steven and Mermelstein, Paul.
\newblock Comparison of parametric representation for monosyllabic word
  recognition in continuously spoken sentences.
\newblock \emph{IEEE Trans. ASSP}, Aug. 1980.

\bibitem[Denil et~al.(2013)Denil, Shakibi, Dinh, de~Freitas,
  et~al.]{denil2013predicting}
Denil, Misha, Shakibi, Babak, Dinh, Laurent, de~Freitas, Nando, et~al.
\newblock Predicting parameters in deep learning.
\newblock In \emph{NIPS}, 2013.

\bibitem[Denton et~al.(2014)Denton, Zaremba, Bruna, LeCun, and
  Fergus]{denton2014exploiting}
Denton, Emily~L, Zaremba, Wojciech, Bruna, Joan, LeCun, Yann, and Fergus, Rob.
\newblock Exploiting linear structure within convolutional networks for
  efficient evaluation.
\newblock In \emph{NIPS}, 2014.

\bibitem[Flanagan(1972)]{Flanagan72}
Flanagan, James~L.
\newblock Speech analysis, synthesis and perception.
\newblock \emph{Springer- Verlag}, 1972.

\bibitem[Gong et~al.(2014)Gong, Liu, Yang, and Bourdev]{gong2014compressing}
Gong, Yunchao, Liu, Liu, Yang, Ming, and Bourdev, Lubomir.
\newblock Compressing deep convolutional networks using vector quantization.
\newblock \emph{arXiv preprint arXiv:1412.6115}, 2014.

\bibitem[Han et~al.(2015)Han, Pool, Tran, and Dally]{han2015learning}
Han, Song, Pool, Jeff, Tran, John, and Dally, William.
\newblock Learning both weights and connections for efficient neural network.
\newblock In \emph{NIPS}, 2015.

\bibitem[Han et~al.(2016)Han, Mao, and Dally]{deep-compression}
Han, Song, Mao, Huizi, and Dally, William~J.
\newblock Deep compression: Compressing deep neural network with pruning,
  trained quantization and huffman coding.
\newblock In \emph{ICLR}, 2016.

\bibitem[He et~al.(2016)He, Zhang, Ren, and Sun]{residual}
He, Kaiming, Zhang, Xiangyu, Ren, Shaoqing, and Sun, Jian.
\newblock Deep residual learning for image recognition.
\newblock In \emph{CVPR}, 2016.

\bibitem[Hinton et~al.(2015)Hinton, Vinyals, and Dean]{dark-knowledge}
Hinton, Geoffrey, Vinyals, Oriol, and Dean, Jeff.
\newblock Distilling the knowledge in a neural network.
\newblock \emph{arXiv preprint arXiv:1503.02531}, 2015.

\bibitem[Howard et~al.(2017)Howard, Zhu, Chen, Kalenichenko, Wang, Weyand,
  Andreetto, and Adam]{mobilenet}
Howard, Andrew~G, Zhu, Menglong, Chen, Bo, Kalenichenko, Dmitry, Wang, Weijun,
  Weyand, Tobias, Andreetto, Marco, and Adam, Hartwig.
\newblock Mobilenets: Efficient convolutional neural networks for mobile vision
  applications.
\newblock \emph{arXiv preprint arXiv:1704.04861}, 2017.

\bibitem[Ioffe \& Szegedy(2015)Ioffe and Szegedy]{batchnorm}
Ioffe, Sergey and Szegedy, Christian.
\newblock Batch normalization: Accelerating deep network training by reducing
  internal covariate shift.
\newblock In \emph{ICML}, 2015.

\bibitem[Jaderberg et~al.(2014)Jaderberg, Vedaldi, and
  Zisserman]{jaderberg2014speeding}
Jaderberg, Max, Vedaldi, Andrea, and Zisserman, Andrew.
\newblock Speeding up convolutional neural networks with low rank expansions.
\newblock \emph{arXiv preprint arXiv:1405.3866}, 2014.

\bibitem[Jin et~al.(2014)Jin, Dundar, and Culurciello]{flattenet}
Jin, Jonghoon, Dundar, Aysegul, and Culurciello, Eugenio.
\newblock Flattened convolutional neural networks for feedforward acceleration.
\newblock \emph{arXiv preprint arXiv:1412.5474}, 2014.

\bibitem[Jin et~al.(2016)Jin, Yuan, Feng, and Yan]{iht}
Jin, Xiaojie, Yuan, Xiaotong, Feng, Jiashi, and Yan, Shuicheng.
\newblock Training skinny deep neural networks with iterative hard thresholding
  methods.
\newblock \emph{arXiv preprint arXiv:1607.05423}, 2016.

\bibitem[Jojic et~al.()Jojic, Frey, and Kannan]{jojic2003epitomic}
Jojic, Nebojsa, Frey, Brendan~J, and Kannan, Anitha.
\newblock Epitomic analysis of appearance and shape.

\bibitem[Kim et~al.(2015)Kim, Park, Yoo, Choi, Yang, and
  Shin]{kim2015compression}
Kim, Yong-Deok, Park, Eunhyeok, Yoo, Sungjoo, Choi, Taelim, Yang, Lu, and Shin,
  Dongjun.
\newblock Compression of deep convolutional neural networks for fast and low
  power mobile applications.
\newblock \emph{arXiv preprint arXiv:1511.06530}, 2015.

\bibitem[Kingma \& Ba(2014)Kingma and Ba]{adam}
Kingma, Diederik and Ba, Jimmy.
\newblock Adam: A method for stochastic optimization.
\newblock In \emph{ICLR}, 2014.

\bibitem[Landone et~al.(2007)Landone, Harrop, and Reiss]{Landone07}
Landone, Christian, Harrop, Joseph, and Reiss, Josh.
\newblock Enabling access to sound archives through integration, enrichment and
  retrieval: the easaier project.
\newblock In \emph{ISMIR}, 2007.

\bibitem[Lebedev \& Lempitsky(2016)Lebedev and Lempitsky]{lebedev2015fast}
Lebedev, Vadim and Lempitsky, Victor.
\newblock Fast convnets using group-wise brain damage.
\newblock In \emph{CVPR}, 2016.

\bibitem[Lebedev et~al.(2014)Lebedev, Ganin, Rakhuba, Oseledets, and
  Lempitsky]{lebedev2014speeding}
Lebedev, Vadim, Ganin, Yaroslav, Rakhuba, Maksim, Oseledets, Ivan, and
  Lempitsky, Victor.
\newblock Speeding-up convolutional neural networks using fine-tuned
  cp-decomposition.
\newblock \emph{arXiv preprint arXiv:1412.6553}, 2014.

\bibitem[Li et~al.(2017)Li, Kadav, Durdanovic, Samet, and Graf]{li2016pruning}
Li, Hao, Kadav, Asim, Durdanovic, Igor, Samet, Hanan, and Graf, Hans~Peter.
\newblock Pruning filters for efficient convnets.
\newblock In \emph{ICLR}, 2017.

\bibitem[Luo et~al.(2017)Luo, Wu, and Lin]{thinet}
Luo, Jian-Hao, Wu, Jianxin, and Lin, Weiyao.
\newblock Thinet: A filter level pruning method for deep neural network
  compression.
\newblock In \emph{ICCV}, 2017.

\bibitem[Mathieu et~al.(2013)Mathieu, Henaff, and LeCun]{mathieu2013fast}
Mathieu, Michael, Henaff, Mikael, and LeCun, Yann.
\newblock Fast training of convolutional networks through ffts.
\newblock \emph{arXiv preprint arXiv:1312.5851}, 2013.

\bibitem[Piczak(2015{\natexlab{a}})]{piczak15}
Piczak, Karol~J.
\newblock Esc: Dataset for environmental sound classification.
\newblock In \emph{ACM MM}, 2015{\natexlab{a}}.

\bibitem[Piczak(2015{\natexlab{b}})]{piczak2015environmental}
Piczak, Karol~J.
\newblock Environmental sound classification with convolutional neural
  networks.
\newblock In \emph{MLSP}, 2015{\natexlab{b}}.

\bibitem[Pols et~al.(1966)]{Pols66}
Pols, Louis~CW et~al.
\newblock Spectral analysis and identification of dutch vowels in monosyllabic
  words.
\newblock \emph{dissertation}, 1966.

\bibitem[Salamon et~al.(2014)Salamon, Jacoby, and Juan~Pable]{Salamon14}
Salamon, Justin, Jacoby, Christopher, and Juan~Pable, Bello.
\newblock A dataset and taxonomy for urban sound research.
\newblock In \emph{ACM MM}, 2014.

\bibitem[Simonyan \& Zisserman(2015)Simonyan and Zisserman]{vgg}
Simonyan, Karen and Zisserman, Andrew.
\newblock Very deep convolutional networks for large-scale image recognition.
\newblock In \emph{ICLR}, 2015.

\bibitem[Sindhwani et~al.(2015)Sindhwani, Sainath, and
  Kumar]{sindhwani2015structured}
Sindhwani, Vikas, Sainath, Tara, and Kumar, Sanjiv.
\newblock Structured transforms for small-footprint deep learning.
\newblock In \emph{NIPS}, 2015.

\bibitem[Srivastava et~al.(2014)Srivastava, Hinton, Krizhevsky, Sutskever, and
  Salakhutdinov]{dropout}
Srivastava, Nitish, Hinton, Geoffrey, Krizhevsky, Alex, Sutskever, Ilya, and
  Salakhutdinov, Ruslan.
\newblock Dropout: A simple way to prevent neural networks from overfitting.
\newblock \emph{JMLR}, 15\penalty0 (1):\penalty0 1929--1958, 2014.

\bibitem[Stowell et~al.(2015)Stowell, Giannoulis, Benetos, Lagrange, and
  Plumbley]{Stowell15}
Stowell, Dan, Giannoulis, Dimitrios, Benetos, Emmanouil, Lagrange, Mathieu, and
  Plumbley, Mark~D.
\newblock Detection and classification of acoustic scenes and events.
\newblock \emph{IEEE Transactions on Multimedia}, 17\penalty0 (10):\penalty0
  1733--1746, 2015.

\bibitem[Szegedy et~al.(2015)Szegedy, Liu, Jia, Sermanet, Reed, Anguelov,
  Erhan, Vanhoucke, and Rabinovich]{googlenet}
Szegedy, Christian, Liu, Wei, Jia, Yangqing, Sermanet, Pierre, Reed, Scott,
  Anguelov, Dragomir, Erhan, Dumitru, Vanhoucke, Vincent, and Rabinovich,
  Andrew.
\newblock Going deeper with convolutions.
\newblock In \emph{CVPR}, 2015.

\bibitem[Xie et~al.(2017)Xie, Girshick, Doll{\'a}r, Tu, and He]{resnext}
Xie, Saining, Girshick, Ross, Doll{\'a}r, Piotr, Tu, Zhuowen, and He, Kaiming.
\newblock Aggregated residual transformations for deep neural networks.
\newblock In \emph{CVPR}, 2017.

\bibitem[Xu et~al.(2008)Xu, Li, and Lee]{Xu08}
Xu, Yangsheng, Li, Wen~Jung, and Lee, Ka~Keung.
\newblock \emph{Intelligent wearable interfaces}.
\newblock John Wiley \& Sons, 2008.

\bibitem[Zhang et~al.(2017)Zhang, Zhou, Lin, and Sun]{shufflenet}
Zhang, Xiangyu, Zhou, Xinyu, Lin, Mengxiao, and Sun, Jian.
\newblock Shufflenet: An extremely efficient convolutional neural network for
  mobile devices.
\newblock \emph{arXiv preprint arXiv:1707.01083}, 2017.

\end{thebibliography}
\bibliographystyle{icml2018}
\clearpage


\section*{Supplementary material}

# WSNet: Compact and Efficient Networks Through Weight Sampling
# — Supplementary Material —

## Abstract

In this supplementary material, we provide detailed experimental settings and more experimental results, including (1) the details of tested datasets; (2) the ablative study of WSNet on MusicDet200K; (3) comparison between WSNet and narrowed baseline on MusicDet200K; (4) the configurations of WSNet used on UrbanSound8K and DCASE; (5) comparison between WSNet and state-of-the-arts on UrbanSound8K and DCASE. (6) the weight quantization method used in experiments; (7) the architecture of baseline network used on CIFAR10.

## 1. Datasets

Details of the four datasets used in our experiments are as follows:

*MusicDet200K* aims to assign a sample a binary label to indicate whether it is music or not. MusicDet200K has overall 238,000 annotated sound clips. Each has a time duration of 4 seconds and is resampled to 16000 Hz and normalized (Piczak, 2015b). Among all samples, we use 200,000/20,000/18,000 as train/val/test set. The samples belonging to "non-music" count for 70% of all samples, which means if we trivially assign all samples to be "non-music", the classification accuracy is 70%.

*ESC-50* (Piczak, 2015a) is a collection of 2000 short (5 seconds) environmental recordings comprising 50 equally balanced classes of sound events in 5 major groups (*animals*, *natural soundscapes and water sounds*, *human non-speech sounds*, *interior/domestic sounds* and *exterior/urban noises*) divided into 5 folds for cross-validation. Following (Aytar et al., 2016), we extract 10 sound clips from each recording with length of 1 second and time step of 0.5 second (*i.e.* two neighboring clips have 0.5 seconds overlapped). Therefore, in each cross-validation, the number of training samples is 16000. In testing, we average over ten clips of each recording for the final classification result.

*UrbanSound8K* (Salamon et al., 2014) is a collection of 8732 short (around 4 seconds) recordings of various urban sound sources (*air conditioner*, *car horn*, *playing children*,

*dog bark*, *drilling*, *engine idling*, *gun shot*, *jackhammer*, *siren* and *street music*). As in ESC-50, we extract 8 clips with the time length of 1 second and time step of 0.5 second from each recording. For those that are less than 1 second, we pad them with zeros and repeat for 8 times (*i.e.* time step is 0.5 second).

*DCASE* (Stowell et al., 2015a) is used in the Detection and Classification of Acoustic Scenes and Events Challenge (DCASE). It contains 10 acoustic scene categories, 10 training examples per category and 100 testing examples. Each sample is a 30-second audio recording. During training, we evenly extract 12 sound clips with time length of 5 seconds and time step of 2.5 seconds from each recording.

## 2. Ablative study of WSNet on MusicDet200K

We conduct extensive ablative study of WSNet on MusicDet200K to investigate the effects of each component of WSNet on final performance. The results are listed in Table 1, which also serve as strong supports to the conclusions made in the ablative analysis (ref. to Sec. 4.2.1) in the main text.

## 3. WSNet versus narrowed baseline on MusicDet200K

As shown in Figure 1, WSNet outperforms baselines by a large magin across all compression ratios. Particularly, when the comparison ratios are large (*e.g.* 42), baselines suffer severe performance drop. In contrast, WSNet achieves comparable accuracies with full-size baselines without significant drop (88.6 versus 88.9). Above results again demonstrate the effectiveness of weight sampling is not due to over-parameterization of baselines.

## 4. The configurations of WSNet on ESC-50, UrbanSound8K and DCASE

Please refer to Table 2.

Table 1: Studies on the effects of different settings of WSNet on the model size, computation cost (in terms of #mult-adds) and classification accuracy on MusicDet200K. Please refer to Table 3 in the main text for the meaning of symbols S/C/D/Q. "SC†" denotes the weight sampling of fully connected layers whose parameters can be seen as flattened vectors with channel of 1. The numbers in subscripts of SC† denotes the compactness of fully connected layers. To avoid confusion, SC† only occured in the names when both spatial and channel sampling are applied for convolutional layers.

| WSNet's settings | conv{1-3} | | | conv4 | | | conv5 | | | conv6 | | | conv7 | | | fc1/2 | Acc. | Model size | Mult-Adds |
|---|---|---|---|---|---|---|---|---|---|---|---|---|---|---|---|---|---|---|---|
| | S | C | D | S | C | D | S | C | D | S | C | D | S | C | D | SC† | | | |
| Baseline | 1 | 1 | 1 | 1 | 1 | 1 | 1 | 1 | 1 | 1 | 1 | 1 | 1 | 1 | 1 | 1 | 88.9 ± 0.1 | 3M (1×) | 1.2e10 (1×) |
| BaselineQ$_4$ | 1 | 1 | 1 | 1 | 1 | 1 | 1 | 1 | 1 | 1 | 1 | 1 | 1 | 1 | 1 | 1 | 88.8 ± 0.1 | 4× | 1× |
| S$_2$ | 2 | 1 | 1 | 2 | 1 | 1 | 2 | 1 | 1 | 2 | 1 | 1 | 2 | 1 | 1 | 2 | 89.0 ± 0.0 | 2× | 1× |
| S$_4$ | 4 | 1 | 1 | 4 | 1 | 1 | 4 | 1 | 1 | 4 | 1 | 1 | 4 | 1 | 1 | 4 | 89.0 ± 0.0 | 4× | 1.8× |
| S$_8$ | 8 | 1 | 1 | 8 | 1 | 1 | 8 | 1 | 1 | 8 | 1 | 1 | 8 | 1 | 1 | 8 | 88.3 ± 0.1 | 5.7× | 3.4× |
| C$_2$ | 1 | 2 | 1 | 1 | 2 | 1 | 1 | 2 | 1 | 1 | 2 | 1 | 1 | 2 | 1 | 2 | 89.1 ± 0.2 | 2× | 1× |
| C$_4$ | 1 | 4 | 1 | 1 | 4 | 1 | 1 | 4 | 1 | 1 | 4 | 1 | 1 | 4 | 1 | 4 | 88.7 ± 0.1 | 4× | 1.4× |
| C$_8$ | 1 | 8 | 1 | 1 | 8 | 1 | 1 | 8 | 1 | 1 | 8 | 1 | 1 | 8 | 1 | 8 | 88.6 ± 0.1 | 8× | 2.4× |
| S$_4$C$_4$SC$^\dagger_4$ | 4 | 4 | 1 | 4 | 4 | 1 | 4 | 4 | 1 | 4 | 4 | 1 | 4 | 4 | 1 | 4 | 88.7 ± 0.0 | 11.1× | 5.7× |
| S$_8$C$_8$SC$^\dagger_8$ | 8 | 8 | 1 | 8 | 8 | 1 | 8 | 8 | 1 | 8 | 8 | 1 | 4 | 8 | 1 | 8 | 88.4 ± 0.0 | 23× | **16.4×** |
| S$_8$C$_8$SC$^\dagger_8$D$_2$ | 8 | 8 | 2 | 8 | 8 | 1 | 8 | 8 | 1 | 8 | 8 | 1 | 4 | 8 | 1 | 8 | **89.2 ± 0.1** | 20× | 3.8× |
| S$_8$C$_8$SC$^\dagger_{15}$D$_2$ | 8 | 8 | 2 | 8 | 8 | 1 | 8 | 8 | 1 | 8 | 8 | 1 | 8 | 8 | 1 | 15 | 88.6 ± 0.0 | 42× | 3.8× |
| S$_8$C$_8$SC$^\dagger_8$Q$_4$ | 8 | 8 | 1 | 8 | 8 | 1 | 8 | 8 | 1 | 8 | 8 | 1 | 4 | 8 | 1 | 8 | 88.4 ± 0.0 | 92× | **16.4×** |
| S$_8$C$_8$SC$^\dagger_{15}$D$_2$Q$_4$ | 8 | 8 | 2 | 8 | 8 | 1 | 8 | 8 | 1 | 8 | 8 | 1 | 8 | 8 | 1 | 15 | 88.5 ± 0.1 | **168×** | 3.8× |

Table 2: The configurations of the WSNet used on UrbanSound8K and DCASE. Please refer to Table 3 in the main text for the meaning of symbols S/C/D/Q. Since the input lengths for the baseline are different in each dataset, we only provide the #Mult-Adds for UrbanSound8K. Note that since we use the ratio of baseline's #Mult-Adds versus WSNet's #Mult-Adds for one WSNet, the numbers corresponding to WSNets in the column of #Mult-Adds are the same for all dataset.

| WSNet's settings | conv{1-4} | | | conv5 | | | conv6 | | | conv7 | | | conv8 | | | Model size | Mult-Adds |
|---|---|---|---|---|---|---|---|---|---|---|---|---|---|---|---|---|---|
| | S | C | D | S | C | D | S | C | D | S | C | D | S | C | D | | |
| Baseline | 1 | 1 | 1 | 1 | 1 | 1 | 1 | 1 | 1 | 1 | 1 | 1 | 1 | 1 | 1 | 13M (1×) | 2.4e9 (1×) |
| S$_8$C$_4$D$_2$ | 4 | 4 | 2 | 4 | 4 | 1 | 4 | 4 | 1 | 4 | 4 | 1 | 8 | 4 | 1 | 25× | 2.3× |
| S$_8$C$_8$D$_2$ | 4 | 4 | 2 | 4 | 4 | 1 | 4 | 4 | 1 | 4 | 8 | 1 | 8 | 8 | 1 | 45× | 2.4× |

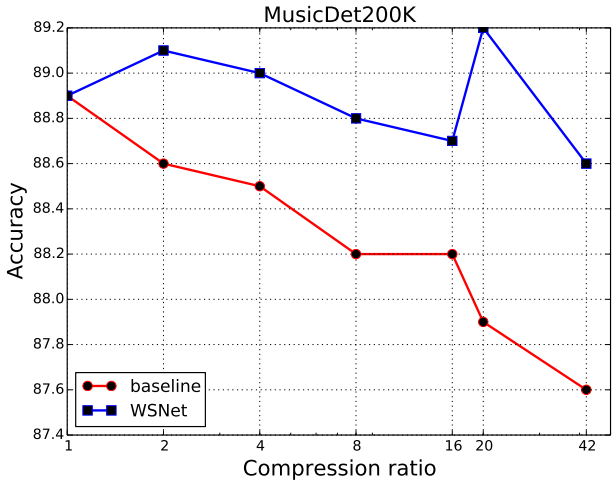

Figure 1: The accuracies of baselines and WSNet with the same model size on MusicDet200K dataset. Note the compression ratios (or compactness for WSNet) are shown in log scale.

Table 3: Comparison with state-of-the-arts using 1D CNNs on UrbanSound8K. All results of WSNet are obtained by 5-folder validation. Please refer to Table 3 in main text for the meaning of symbols S/C/D/Q. Note that Piczak ConvNet (Piczak, 2015b) uses pre-computed 2D frequency images as input, while others use raw audio wave as input.

| Model | Acc. (%) | Model size |
|---|---|---|
| baseline | $70.39 \pm 0.31$ | 13M (1×) |
| baseline$Q_4$ | $70.10 \pm 0.31$ | 4× |
| WSNet (S$_8$C$_4$D$_2$) | $70.76 \pm 0.15$ | 25× |
| WSNet (S$_8$C$_4$D$_2$Q$_4$) | $\mathbf{70.61 \pm 0.20}$ | **100×** |
| WSNet (S$_8$C$_8$D$_2$) | $70.14 \pm 0.23$ | 45× |
| WSNet (S$_8$C$_8$D$_2$Q$_4$) | $\mathbf{70.03 \pm 0.11}$ | **180×** |
| Piczak ConvNet (Piczak, 2015b) | 73.1 | 28M |

## 5. Comparison with state-of-the-art on UrbanSound8K and DCASE

### 5.1. UrbanSound8K

We report the comparison results of WSNet with state-of-the-arts on UrbanSound8k in Table 3. It is observed that WS-Net significantly reduces the model size of baseline while obtaining comparative results. Both (Piczak, 2015b) and (Salamon & Bello, 2015) use pre-computed 2D frequency features after log-mel transformation as input. In comparison, the proposed WSNet simply takes the raw wave of recordings as input, enabling the model to be trained in an end-to-end manner.

Table 4: Comparison with state-of-the-arts using 1D CNNs on DCASE. Note there are only 100 samples in testing set. Please refer to Table 3 in main text for the meaning of symbols S/C/D/Q. Note SoundNet* uses extra data during training while others only use provided training data.

| Model | Acc. (%) | Model size |
|---|---|---|
| baseline | $85 \pm 0$ | 13M (1×) |
| baseline$Q_4$ | $84 \pm 0$ | 4× |
| WSNet (S$_8$C$_4$D$_2$) | $86 \pm 0$ | 25× |
| WSNet (S$_8$C$_4$D$_2$Q$_4$) | $\mathbf{86 \pm 0}$ | **100×** |
| WSNet (S$_8$C$_8$D$_2$) | $84 \pm 0$ | 45× |
| WSNet (S$_8$C$_8$D$_2$Q$_4$) | $\mathbf{84 \pm 0}$ | **180×** |
| RNH (Roma et al., 2013) | 77 | - |
| Ensemble (Stowell et al., 2015b) | 78 | - |
| SoundNet* (Aytar et al., 2016) | 88 | 13M |

### 5.2. DCASE

As evidenced in Table 4, WSNet outperforms the classification accuracy of the baseline by 1% with a 100× smaller model. When using an even more compact model, *i.e.* 180× smaller in model size. The classification accuracy of WS-Net is only one percentage lower than the baseline (*i.e.* has only one more incorrectly classified sample), verifying the effectiveness of WSNet in learning discriminative representatiosn with highly efficient network. Compared with SoundNet (Aytar et al., 2016) that utilizes a large number of unlabeled data during training, WSNet (S$_8$C$_4$D$_2$Q$_4$) that is 100× smaller achieves comparable results only by using the provided data.

## 6. Weight quantization

Similar to other works (Han et al., 2016; Rastegari et al., 2016), we apply weight quantization to further reduce the size of WSNet. Specifically, the weights in each layer are linearly quantized to $q$ bins where $q$ is a pre-defined number. By setting all weights in the same bin to the same value, we only need to store a small index of the shared weight for each weight. The size of each bin is calculated as $(\max(\mathbf{\Phi}) - \min(\mathbf{\Phi}))/q$. Given $q$ bins, we only need $\log_2(q)$ bits to encode the index. Assuming each weight in WSNet is represented using 4 bytes float number (32 bits) without weight quantization, the ratio of each layer's size before and after weight quantization is $\frac{32 L^* M^*}{L^* M^* \cdot \log_2(q) + 32q}$. Recall that $L^*$ and $M^*$ are the spatial size and the channel number of condensed filter. Since the condition $L^* M^* \gg q$ generally holds in most layers of WSNet, weight quantization is able to reduce the model size by a factor of $\frac{32}{\log_2(q)}$. Different from (Han et al., 2016; Rastegari et al., 2016) which learns the quantization during training, we apply weight quantization to WSNet after its training. In the experiments, we find that such an off-line way is sufficient to reduce model size

Table 5: Configurations of the baseline network (Chen et al., 2016) used on CIFAR10. Each convolutional layer is followed by a nonlinearity layer (*i.e.* ReLU). There are max-pooling layers (with size of 2 and stride of 2) and drop out layers following conv2, conv4 and conv5. The nonlinearity layers, max-pooling layers and dropout layers are omitted in the table for brevity. The padding strategies are all "size preserving".

| Layer | conv1 | conv2 | conv3 | conv4 | conv5 | fc1 |
|---|---|---|---|---|---|---|
| Filter sizes | 5×5 | 5×5 | 5×5 | 5×5 | 5×5 | 4096 |
| #Filters | 32 | 64 | 64 | 128 | 256 | 10 |
| Stride | 1 | 1 | 1 | 1 | 1 | 1 |
| #Params | 2K | 51K | 102K | 205K | 819K | 40K |

without losing accuracy.

# 7. The baseline nework used on CIFAR10

Please refer to Table 5.

# References

Aytar, Yusuf, Vondrick, Carl, and Torralba, Antonio. Soundnet: Learning sound representations from unlabeled video. In *NIPS*, 2016.

Chen, Wenlin, Wilson, James T, Tyree, Stephen, Weinberger, Kilian Q, and Chen, Yixin. Compressing convolutional neural networks in the frequency domain. In *KDD*, 2016.

Han, Song, Mao, Huizi, and Dally, William J. Deep compression: Compressing deep neural network with pruning, trained quantization and huffman coding. In *ICLR*, 2016.

Piczak, Karol J. Esc: Dataset for environmental sound classification. In *ACM MM*, 2015a.

Piczak, Karol J. Environmental sound classification with convolutional neural networks. In *MLSP*, 2015b.

Rastegari, Mohammad, Ordonez, Vicente, Redmon, Joseph, and Farhadi, Ali. Xnor-net: Imagenet classification using binary convolutional neural networks. In *ECCV*, 2016.

Roma, Gerard, Nogueira, Waldo, Herrera, Perfecto, and de Boronat, Roc. Recurrence quantification analysis features for auditory scene classification. *IEEE AASP Challenge on Detection and Classification of Acoustic Scenes and Events*, 2, 2013.

Salamon, Justin and Bello, Juan Pablo. Unsupervised feature learning for urban sound classification. In *ICASSP*, 2015.

Salamon, Justin, Jacoby, Christopher, and Juan Pable, Bello. A dataset and taxonomy for urban sound research. In *ACM MM*, 2014.

Stowell, Dan, Giannoulis, Dimitrios, Benetos, Emmanouil, Lagrange, Mathieu, and Plumbley, Mark D. Detection and classification of acoustic scenes and events. *IEEE Transactions on Multimedia*, 17(10):1733–1746, 2015a.

Stowell, Dan, Giannoulis, Dimitrios, Benetos, Emmanouil, Lagrange, Mathieu, and Plumbley, Mark D. Detection and classification of acoustic scenes and events. *IEEE Transactions on Multimedia*, 17(10):1733–1746, 2015b.


\end{document}